\documentclass[acmconf,screen]{acmart}
\AtBeginDocument{%
  \providecommand\BibTeX{{%
    \normalfont B\kern-0.5em{\scshape i\kern-0.25em b}\kern-0.8em\TeX}}}

\setcopyright{none}
\usepackage{tikz}
\usetikzlibrary{arrows,shapes}
\usetikzlibrary{trees}
\usetikzlibrary{matrix,arrows} 				
\usetikzlibrary{positioning}				
\usetikzlibrary{calc,through}				
\usetikzlibrary{decorations.pathreplacing}  
\usepackage{pgffor}	
\usetikzlibrary{decorations.pathmorphing}	
\usetikzlibrary{decorations.markings}
\usetikzlibrary{snakes}
\tikzset{
    vector/.style={decorate, decoration={snake}, draw},
	provector/.style={decorate, decoration={snake,amplitude=2.5pt}, draw},
	antivector/.style={decorate, decoration={snake,amplitude=-2.5pt}, draw},
    fermion/.style={draw=black, postaction={decorate},
        decoration={markings,mark=at position .55 with {\arrow[draw=black]{>}}}},
    fermionbar/.style={draw=black, postaction={decorate},
        decoration={markings,mark=at position .55 with {\arrow[draw=black]{<}}}},
    fermionnoarrow/.style={draw=black},
    gluon/.style={decorate, draw=black,
        decoration={coil,amplitude=3pt, segment length=5pt}},
    scalar/.style={dashed,draw=black, postaction={decorate},
        decoration={markings,mark=at position .55 with {\arrow[draw=black]{>}}}},
    scalarbar/.style={dashed,draw=black, postaction={decorate},
        decoration={markings,mark=at position .55 with {\arrow[draw=black]{<}}}},
    scalarnoarrow/.style={dashed,draw=black},
    electron/.style={draw=black, postaction={decorate},
        decoration={markings,mark=at position .55 with {\arrow[draw=black]{>}}}},
	bigvector/.style={decorate, decoration={snake,amplitude=4pt}, draw},
}

\tikzstyle{block} = [draw, rectangle, 
    minimum height=3em, minimum width=6em]

\acmConference[ANDEA '21]{ANDEA '21: Anomaly and Novelty Detection, Explanation and Accommodation}
\acmBooktitle{Anomaly and Novelty Detection, Explanation and Accommodation}

\begin{document}

\title[New Methods and Datasets for Group Anomaly Detection]{New Methods and Datasets for Group Anomaly Detection From Fundamental Physics}

\author{Gregor Kasieczka}
\email{gregor.kasieczka@uni-hamburg.de}
\orcid{0000-0003-3457-2755}
\affiliation{%
  \institution{Institut f\"ur Experimentalphysik, Universit\"at Hamburg}
  \streetaddress{Luruper Chaussee 149}
  \city{Hamburg}
  \country{Germany}
  \postcode{D-22761}
}
\author{Benjamin Nachman}
\email{bpnachman@lbl.gov}
\orcid{0000-0003-1024-0932}
\affiliation{%
  \institution{Physics Division, Lawrence Berkeley National Laboratory}
  \streetaddress{???}
  \city{Berkeley}
  \country{CA, USA}
  \postcode{???}
}
\author{David Shih}
\email{shih@physics.rutgers.edu}
\orcid{0000-0003-3408-3871}
\affiliation{%
  \institution{NHETC, Department of Physics \& Astronomy, Rutgers University}
  \streetaddress{???}
  \city{Piscataway}
  \country{NJ, USA}
  \postcode{08854}
}

\renewcommand{\shortauthors}{Kasieczka, Nachman, and Shih}

\begin{abstract}
The identification of anomalous overdensities in data --- group or collective anomaly detection --- is a rich problem with a large number of real world applications. However, it has received relatively little attention in the broader ML community, as compared to point anomalies or other types of single instance outliers. One reason for this is the lack of powerful benchmark datasets.  In this paper, we first explain how, after the Nobel-prize winning discovery of the Higgs boson, unsupervised group anomaly detection has become a new frontier of fundamental physics (where the motivation is to find new particles and forces).
Then we propose a realistic synthetic benchmark dataset (LHCO2020)
for the development of group anomaly detection algorithms. Finally, we compare several existing statistically-sound techniques for unsupervised group anomaly detection, and demonstrate their performance on the LHCO2020 dataset. 
\end{abstract}

\ccsdesc[500]{Computing methodologies~Anomaly detection}
\ccsdesc[500]{Applied computing~Physics}

\keywords{anomaly detection, outlier detection, group anomalies, benchmark dataset, density estimation}


\maketitle

\section{Introduction}

Unsupervised anomaly detection is a long established area of statistics~\cite{grubbs1969procedures} and has recently seen substantial progress from modern machine learning approaches (see Refs.~\cite{chalapathy2019deep,kwon2019survey} for recent reviews).
Most of these methods are designed to identify examples that are individually anomalous, i.e. $\Pr(\mathrm{example}|\mathrm{normal})$ is vanishingly small.  An area of anomaly detection that has received comparatively less attention is the case where one cannot determine with certainty that any single example is anomalous.  These ``group anomalies" instead manifest as overdensities in the probability density of the data and occur naturally in a variety of applications ranging from computer security to pandemic detection and include many other scientific, industrial, and financial applications.

Group anomaly detection is also central to fundamental physics.  In particular, most data analyses at the Large Hadron Collider (LHC) can be seen as group anomaly detection.  Of these approaches, nearly all of them are supervised and rely strongly on a particular anomaly hypothesis.  Following the Nobel-prize winning discovery of the Higgs boson in 2012 (which was a benchmark for group anomaly detection~\cite{muandet2013oneclass}), there is increasing urgency in searching for new phenomena beyond the Standard Model (BSM) of particle physics, for which there is ample indirect evidence (e.g.~dark matter). Since these new BSM particles and interactions can take nearly any form, there is correspondingly a growing need for unsupervised approaches to group anomaly detection at the LHC. 

As we will review in section \ref{subsec:related}, existing approaches to group anomaly detection in the machine learning literature are insufficiently sensitive for applying to fundamental physics or operate under a different set of assumptions. This has inspired researchers in fundamental physics to devise a host of new approaches to unsupervised group anomaly detection. In this paper, our first goal is to clearly define the challenge of modern group anomaly detection at the LHC and how it relates to other applications (Sec.~\ref{sec:problem}), and to highlight a few especially promising approaches that have been recently proposed in the LHC literature. Group anomaly detection at the LHC shares many properties with other possible applications, so our expectation is that results obtained there will be useful more widely. 

The second purpose of this paper is to bring the LHC Olympics  2020 (LHCO2020) challenge and datasets~\cite{2101.08320} to the attention of the wider ML community. The LHCO2020 was initiated in order to facilitate the development and comparison of semi-, weakly-, and un-supervised (which we call less-than-supervised) group anomaly detection methods. In this paper, we propose the LHCO2020 datasets as high quality findable, accessible, interoperable, and reusable (FAIR) benchmarks for group anomaly detection in a broader setting.  The datasets have been well-curated and documented by domain experts, but can be used by others without specific domain knowledge. In Sec.~\ref{sec:data}, we will describe the datasets in more detail. In Sec.~\ref{sec:methods}, we will briefly review some machine learning methods that have already been developed and applied to the LHCO2020 dataset, and in Sec.~\ref{sec:experiments} we will provide some quantitative comparisons of selected methods.  

\subsection{Problem statement}
\label{sec:problem}

Data at the LHC are obtained by smashing beams of protons at relativistic velocities.  The large energy $E$ available in these collisions allows the creation of new particles with mass $m$ via $E=mc^2$, where $c$ is the speed of light.  Such particles then decay and radiate, producing a spray of secondary particles that interact with the detector and register signals in millions of readout channels.  Standard dimensionality reduction schemes are then used to reconstruct the secondary particles.  Further (lossy) dimensionality reduction is accomplished by combining these trajectories using physics-inspired functions.  A typical data analysis will use features $x\in{\mathbb R}^N$ for $N\sim 10$.

One aspect of data analysis at the LHC that distinguishes it from other areas is the excellent fidelity of available simulators. The typical analysis workflow at the LHC begins with a particular anomaly hypothesis.  
Probability densities $p(x|\mathrm{anomaly})$ and $p(x|\mathrm{normal})$ 
are then numerically estimated using simulators.
Finally, a likelihood ratio test with the empirical data probability density $p(x|\mathrm{data})$ is used to test for the presence of anomalous events.

While this procedure has achieved broad sensitivity to a variety of anomaly hypotheses, it also has large gaps in coverage.  In particular, not all collision types can be accurately simulated and not all possible anomaly hypotheses are known.  It is therefore essential to use unsupervised approaches that do not rely on positing a particular anomaly hypothesis and can estimate likelihood ratios directly from unlabeled data.  

There is no known general solution to this challenge, but one class of group anomalies in collider physics called \textit{resonances} are amenable to less-than-supervised learning.  These anomalies have the following generic characteristics:

\begin{itemize}
    \item Rarity: $\Pr(\mathrm{anomaly})\ll \Pr(\mathrm{normal})$.
    \item Overlap: $\max_x p(x|\mathrm{anomaly})/p(x|\mathrm{normal})<\infty$.
    \item Resonance: $\Pr(|m-m_0| >\delta |\mathrm{anomaly})\approx 0$ for some feature $m$ (which is often a mass) and fixed $m_0,\delta$.
    \item Smoothness: $p(x|m,\mathrm{normal})$ varies slowly with $m$ so that one can use data with $|m-m_0|>\delta$ to estimate $p(x|m,\mathrm{normal})$ for $|m-m_0| < \delta$.
\end{itemize}

 Apart from the above, no assumptions on the anomaly are made. Specifically there is no preferred value of $m_0$ or which features are sensitive to it. Furthermore, no group memberships are known.

\textit{Rarity} is required because otherwise the anomalies would have been ruled out by non-observation in existing analyses.  In the LHC Olympics, the anomalies constitute much less than 1\% of the dataset.  The \textit{overlapping} support distinguishes these group anomalies from off-manifold anomalies and occurs physically due to radiative processes and detector effects that ensure that anything that can happen, will happen with some probability.  \textit{Resonance} is natural for searches looking for a particle of mass $m_0$ that decays into objects which can be fully observed.  The reconstructed mass $m$ of the decay products will naturally be localized near $m_0$, where the spread $\delta$ is often dominated by detector effects and independent of the details of the new particle.  \textit{Smoothness} is an excellent approximation because the physical processes underlying the particle interactions do not change abruptly as a function of $m$.


Group anomaly detection in other domains have similar properties as resonant anomalies in collider physics.  An example is the detection of distributed denial of service (DDoS) attacks on computer networks: 
In this case, the feature encoding time takes the role of the invariant mass in physics problems.  The number of malicious packages is small compared to overall traffic (\textit{Rarity}), individual packages look innocent (\textit{Overlap}), but share a set of properties such as used protocol or originating hosts. Finally these attacks
are limited in time (\textit{Resonance}) over the stationary/periodic background of network traffic (\textit{Smoothness}).

Similar correspondences exist for other challenges in fundamental science (e.g. galaxy classification), 
engineering (e.g. predictive maintenance or production line monitoring),
medicine (e.g. early warning systems for pandemics or cancer detection),
social media analysis (e.g. trending tweets)
and financial data analysis (e.g. insurance fraud, credit card fraud or stock market analysis). 

\subsection{Approaches}

The proposed methods for group anomaly detection rely on defining a \emph{signal region} (SR) --- a compact
space defined solely in $m$ so that in it, the data is possibly enriched in signal.
The complement is termed \emph{sideband} (SB).  At the LHC, a common choice for the SR is an interval in $m$ centered around $m_0$ with a width related to $\delta$.  However, in general the position $m_0$ of a potential signal is not known. This can be solved with a sliding window approach for the SR. 
Results obtained using different regions can be used individually (no group anomaly in this region)
or statistically combined --- by computing appropriate trial factors --- for a global statement (presence or absence of a group anomaly in the data).
This method is also often referred to as \textit{bump hunting}. In particle 
physics specifically, bump hunting has a long history~\cite{PhysRevLett.33.1404,PhysRevLett.33.1406}, but the \textit{Rarity} assumption in this problem requires additional statistical methods
to enhance infrequent anomalies in an unsupervised way.


We will introduce two algorithms for group anomaly detection: classification-based and density-based. In the first case, we construct a binary classifier to distinguish between the SR and SB using features that are independent from $m$. Simply put: if non-trivial classification can achieved, it is a sign for an underlying difference between the sideband and signal region --- hence a group anomaly. The second approach directly estimates the density of normal events conditional on $m$ in the SB and then interpolates it to the SR where it can be used for likelihood-ratio based anomaly detection. A detailed description of the algorithms is provided in Sec.~\ref{sec:methods} and code is available from the authors upon request.

\subsection{Related work}
\label{subsec:related}

Alternate approaches for group anomaly detection 
fall into two categories~\cite{toth2018group}: discriminative and generative.

Two classic examples of discriminative methods are One-Class Support Machines (OCSMM)~\cite{muandet2013oneclass} and Support Measure Data Description (SMDD)~\cite{guevara:hal-01330487}. Both assume known group memberships and test whether a given group is anomalous with respect to a distribution of normal groups.
In contrast, we address the problem when no group membership is known for individual points. Imposing groups via clustering is not feasible as the anomalous group overlaps with the normal group and only differs in density.

For the same reason, generative approaches that require known group memberships~\cite{xiong2011group,xiong2011hierarchical,chalapathy2018group} cannot be applied. Group Latent Anomaly Detection (GLAD) is a one-step model that unifies
group discovery and anomaly detection~\cite{yu2015}, developed for social media analysis. A related approach based on topic modelling and Latent Dirichlet Allocation (LDA)~\cite{{Dillon:2019cqt,Dillon:2020quc}} was deployed on the LHCO2020 datasets~\cite{2101.08320}. However, other methods --- such as density estimation based anomaly detection reviewed in this contribution --- outperform LDA on the LHCO2020 dataset.

An interesting connection exists to developments in point based
anomaly detection using density estimation. Recently~\cite{lan2020perfect} showed that low background likelihood by itself is not a reliable anomaly metric and a likelihood ratio is needed instead.
Constructing such a likelihood ratio is often difficult  and 
ad-hoc background models based on 
perturbing the data~\cite{ren2019likelihood},
measuring input complexity~\cite{serra2020input},
or a second density estimator trained on generic data~\cite{schirrmeister2020understanding}
are used in in practice. Compared to these,
the \textit{Resonance} and \textit{Smoothness} assumptions allow
robust construction of a likelihood ratio.

The absence of realistic open benchmark datasets for development and evaluation is a known problem limiting progress in methods for anomaly detection~\cite{toth2018group,Pang2020DeepLF}.
Purely synthetic datasets --- such as Gaussian distributions with different correlations or mixed examples e.g. from different image classes~\cite{xiong2011group,chalapathy2019deep} --- have limited realism and practical applicability.  
Other datasets from the natural sciences~\cite{xiong2011group,xiong2011hierarchical,muandet2013oneclass,guevara:hal-01330487} are either not available at all or require substantial amounts of further processing before being suitable for anomaly detection studies.
Similarly, datasets from network intrusion detection~\cite{nlskdd,Divekar_2018} in general require domain knowledge and feature engineering and are --- depending on the time slice --- too densely populated with anomalies.

Compared to other available datasets~\cite{Rayana:2016}, LHCO2020 is ready-to-use without 
additional pre-processing or feature engineering required. The feature representations at two different levels of complexity allow developing end-to-end algorithms as well as fast prototyping on low-dimensional features.
Injected anomalies are rare 
(e.g., 834 anomalies examples out of 1M data points for black box 1)
and have high point difficulty (all anomalies are inliers)
but have sufficient complexity (up to 2100 features/examples)
to allow succesful detection.
Although based on simulation tools, consistent 
methods are employed to simulate background and anomaly and the overall toolchain is well-validated including comparisons to experimental data. For added realism, different background models are used in the development datasets and the individual challenge sets (see the following Section for details). Finally, the LHCO2020 dataset was extensively vetted by domain experts and a number of anomaly detection algorithms were evaluated on it\footnote{After submission of this work, a second anomaly detection challenge aimed at finding anomalies in particle physics data was published~\cite{dankMachines}, further underscoring the relevance of this task.}.

\section{LHC Olympics Dataset}
\label{sec:data}

The portal for the LHC Olympics datasets can be found at the challenge website\footnote{\url{https://lhco2020.github.io/homepage/}}.  The datasets described below are all publicly available and downloadable from Zenodo~\cite{lhc_bb1}. The LHCO2020 consist of a dataset 
for algorithm development (R\&D Dataset) and three initially blinded datasets (Black Box 1--3). Following
conclusion of the challenge phase, labels (anomaly or not) are now provided for these datasets as well.

Standard Monte Carlo based tools were used to create these datasets. The underlying physics process was simulated via \textsc{Pythia}~\cite{Sjostrand:2007gs} and \textsc{Herwig++}~\cite{Bahr:2008pv} while \textsc{Delphes}~\cite{deFavereau:2013fsa}
was used to model the finite detector resolution.  These software packages are highly configurable via so-called \textit{tunes} which encode empirical properties of fundamental physics and uncertainties. 
To increase the realism of the challenge, different tunes were used to create the datasets as described in the following.
For more details on the (public, open source) tools that were used to produce these simulated events, we refer the reader to the challenge website.

All datasets are arrays (pandas dataframes~\cite{mckinney-proc-scipy-2010} saved to compressed HDF5~\cite{koranne2011hierarchical} format) with shape ($N_\text{events}$, 2101). 
Each row of the array is a single data instance (event), representing the products of a single collision in the simulated particle detector. 

%


\subsection{R\&D Dataset}
\label{sec:challenge_rnd}

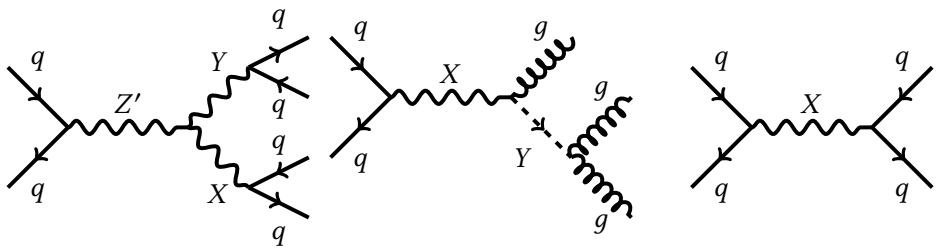
\begin{figure}[h!]

\begin{minipage}{6in}
  \centering
\raisebox{-0.5\height}{
\begin{tikzpicture}[line width=1.5 pt,scale=0.8]
	\draw[fermion] (-2,1) -- (-1,0);
	\draw[fermion] (-1,0) -- (-2,-1);
	\draw[vector] (-1,0) -- (1,0);
	\draw[vector] (2,1) -- (1,0);
	\draw[vector] (1,0) -- (2,-1);
	\draw[fermion] (2,1) -- (3,1.5);
	\draw[fermion] (3,0.5) -- (2,1);
	\draw[fermion] (2,-1) -- (3,-1.5);
	\draw[fermion] (3,-0.5) -- (2,-1);
	\node at (-1.5, -1.1) {\Large $q$};
	\node at (-1.5, 1.1) {\Large $q$};
	\node at (1.5, -1.1) {\Large $X$};
	\node at (1.5, 1.1) {\Large $Y$};
	\node at (2.5, 0.3) {\Large $q$};
	\node at (2.5, 1.8) {\Large $q$};
	\node at (2.5, -0.3) {\Large $q$};
	\node at (2.5, -1.8) {\Large $q$};
	\node at (0,0.4) {\Large $Z'$};
 \end{tikzpicture}
 }
  \raisebox{-0.5\height}{
  \begin{tikzpicture}[line width=1.5 pt,scale=0.8]
	\draw[fermion] (-2,1) -- (-1,0);
	\draw[fermion] (-1,0) -- (-2,-1);
	\draw[vector] (-1,0) -- (1,0);
	\draw[gluon] (2,1) -- (1,0);
	\draw[scalar] (1,0) -- (2,-1);
	\draw[gluon] (2,-1) -- (3,-2);
	\draw[gluon] (2,-1) -- (3,0);
	\node at (-1.5, -1.1) {\Large $q$};
	\node at (-1.5, 1.1) {\Large $q$};
	\node at (1.2, -1.) {\Large $Y$};
	\node at (1.5, 1.1) {\Large $g$};
	\node at (2.5, .1) {\Large $g$};
	\node at (2.5, -2.1) {\Large $g$};
	\node at (0,0.4) {\Large $X$};
\end{tikzpicture}}
  \hspace*{.2in}
  \raisebox{-0.5\height}{
  \begin{tikzpicture}[line width=1.5 pt,scale=0.8]
	\draw[fermion] (-2,1) -- (-1,0);
	\draw[fermion] (-1,0) -- (-2,-1);
	\draw[vector] (-1,0) -- (1,0);
	\draw[fermion] (2,1) -- (1,0);
	\draw[fermion] (1,0) -- (2,-1);
	\node at (-1.5, -1.1) {\Large $q$};
	\node at (-1.5, 1.1) {\Large $q$};
	\node at (1.5, -1.1) {\Large $q$};
	\node at (1.5, 1.1) {\Large $q$};
	\node at (0,0.4) {\Large $X$};
 \end{tikzpicture}
  }
\end{minipage}

\caption{Schematic diagrams (Feynman diagram) of the anomaly used for 
the R\&D dataset and Black Box 1 (left) as well as 
tri-jet (center) and di-jet (right) anomalies used for Black Box 3 . Incoming particles from collisions are on the left,
outgoing particles  eventually measured by detectors are on the right. 
Lines in the middle represent virtual resonances leading to anomalous
correlations of features in observed events. Different linestyles denote different types of particles.
}
\label{fig:bb1sig}
\end{figure}

The events in the R\&D dataset are of two types. The ``normal" or ``background" events consist of one million 
events produced through the strong interactions of the Standard Model. The ``anomaly" or ``signal" events consist of 100,000 events where a hypothetical heavy particle 
from a theory Beyond the Standard Model (BSM) (called the $Z'$) decays instantaneously to two additional heavy BSM particles (called $X$ and $Y$), which each decay to a collimated spray of Standard Model particles (hadrons). See Fig.~\ref{fig:bb1sig} for an illustration of the anomaly events.  

\begin{figure}[t!]

\begin{minipage}{6in}
  \centering
  \raisebox{-0.5\height}{\includegraphics[width=0.27\columnwidth]{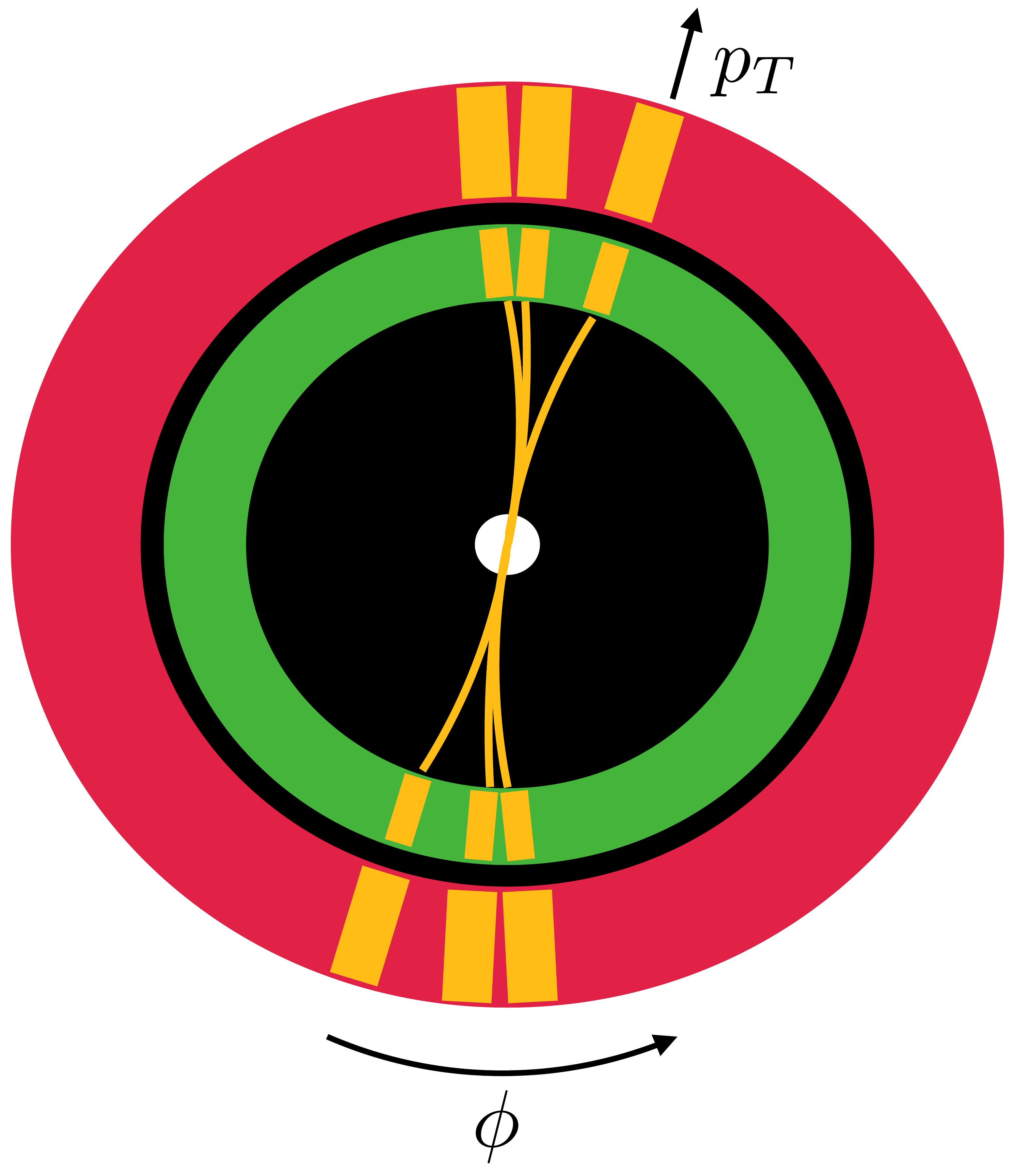}}
  \hspace*{.2in}
  \raisebox{-0.5\height}{\includegraphics[width=0.27\columnwidth]{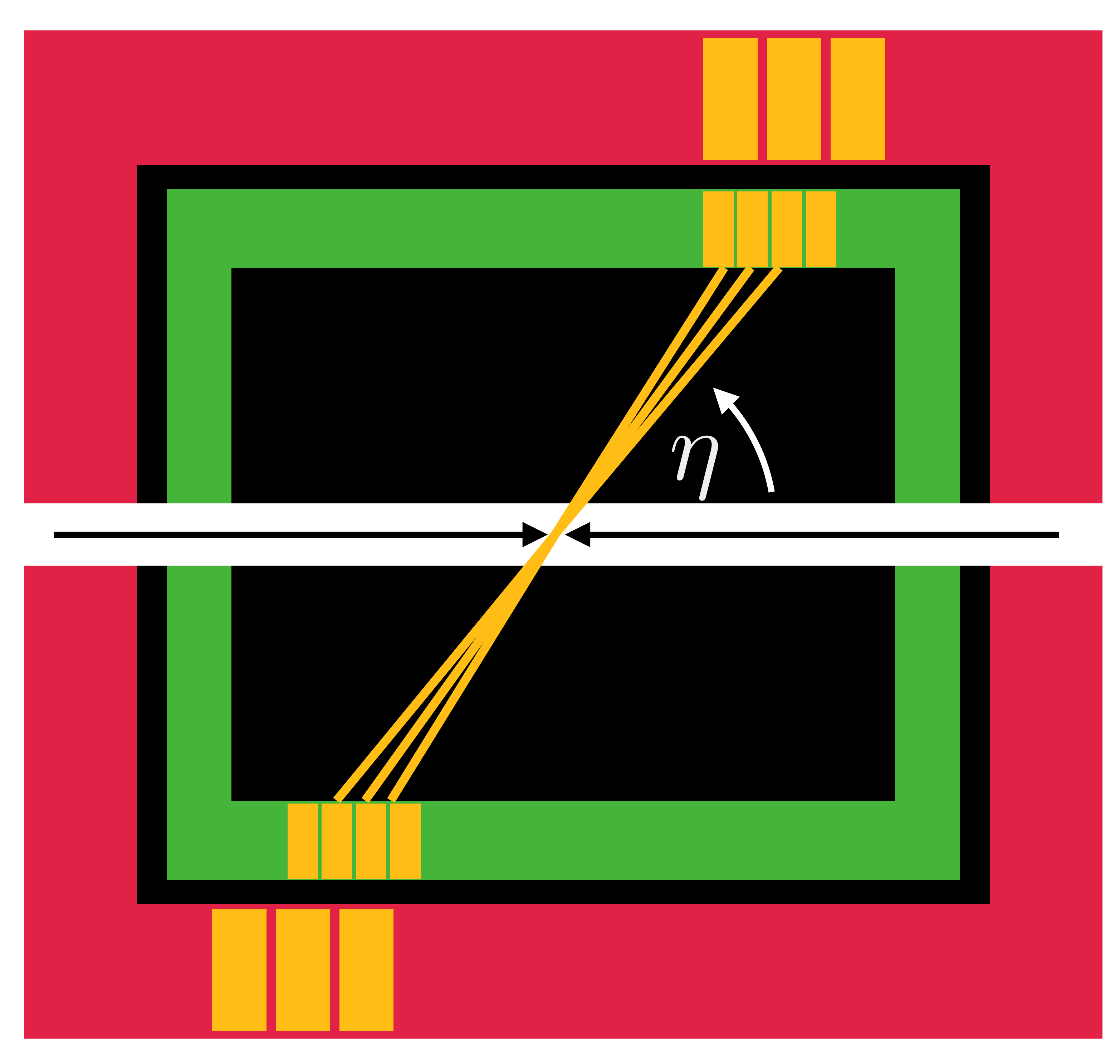}}
\end{minipage}

\caption{A schematic diagram of a detector at the LHC to illustrate the standard coordinate system.  In the top view, protons collide into and out of the page while in the bottom view, protons collide from the left and right.  The collision debris flies out in all directions and for simplicity is represented by six particles.  These particles register signals in a series of detector components.  Their trajectories are then reconstructed using their transverse momentum $p_T$ and angular coordinates $\phi$ and $\eta$. }

\label{fig:geo}
\end{figure}

In each event (background or signal), the properties of the 700 most energetic collision products (particles known as ``hadrons") are recorded in standard particle physics detector coordinates, $p_\text{T}$ -- ``transverse momentum", $\eta$ -- ``pseudorapidity" and $\phi$ -- ``azimuthal angle" as illustrated in Fig.~\ref{fig:geo}. More detailed information such as particle charge is not included. If the event has fewer than 700 collision products, it is zero padded. Finally, the truth bit (signal or background) is appended at the end of every event. In this way, each event 
comprises 2101 floating point numbers.  

The purpose of the R\&D dataset is to provide a common benchmark dataset for anomaly detection techniques on a realistic and well-motivated BSM signal.\footnote{The signal model was discussed in Ref.~\cite{1907.06659} and has the feature that existing searches for BSM physics  may not be particularly sensitive.
} For this purpose, many more anomaly events (100k) were provided than would be present in a realistic setting (e.g., 1000 out of 1M background events). Also, the truth bit labels were provided so that anomaly detection approaches could be evaluated, whereas in real data the truth bit is not known --- hence the need for group anomaly detection.

Finally, in addition to the raw features dataset described above, a reduced dataset of high-level features was also made available during the course of the challenge. These high-level features are computed from those of the raw dataset, but they are provided separately for convenience. They are summaries of the overall geometrical arrangement and energy distribution of the low-level features.

Concretely, the particles are first grouped using a hierarchical clustering algorithm~\cite{Cacciari:2008gp} to form so-called \textit{jets} --- collimated sprays of hadrons. The two most energetic jets in each event are selected and the remaining jets are discarded. Then the relativistic invariant mass of these two jets is calculated. This is the potentially resonant feature $m$.
The other high-level features are: $m_{j_1}$ the invariant mass of the lighter jet;
$\Delta m_j$ the mass difference of the two jets;
and $\tau_{21,1}$ and $\tau_{21,2}$ the $N$-subjettiess ratios~\cite{Thaler:2010tr,Thaler:2011gf} of the leading two jets. 
This feature quantifies the degree to which a jet is characterized by two subjets or one subjet, with smaller values indicating two-prong substructure.

\begin{figure}[h!]
\includegraphics[width=0.47\columnwidth]{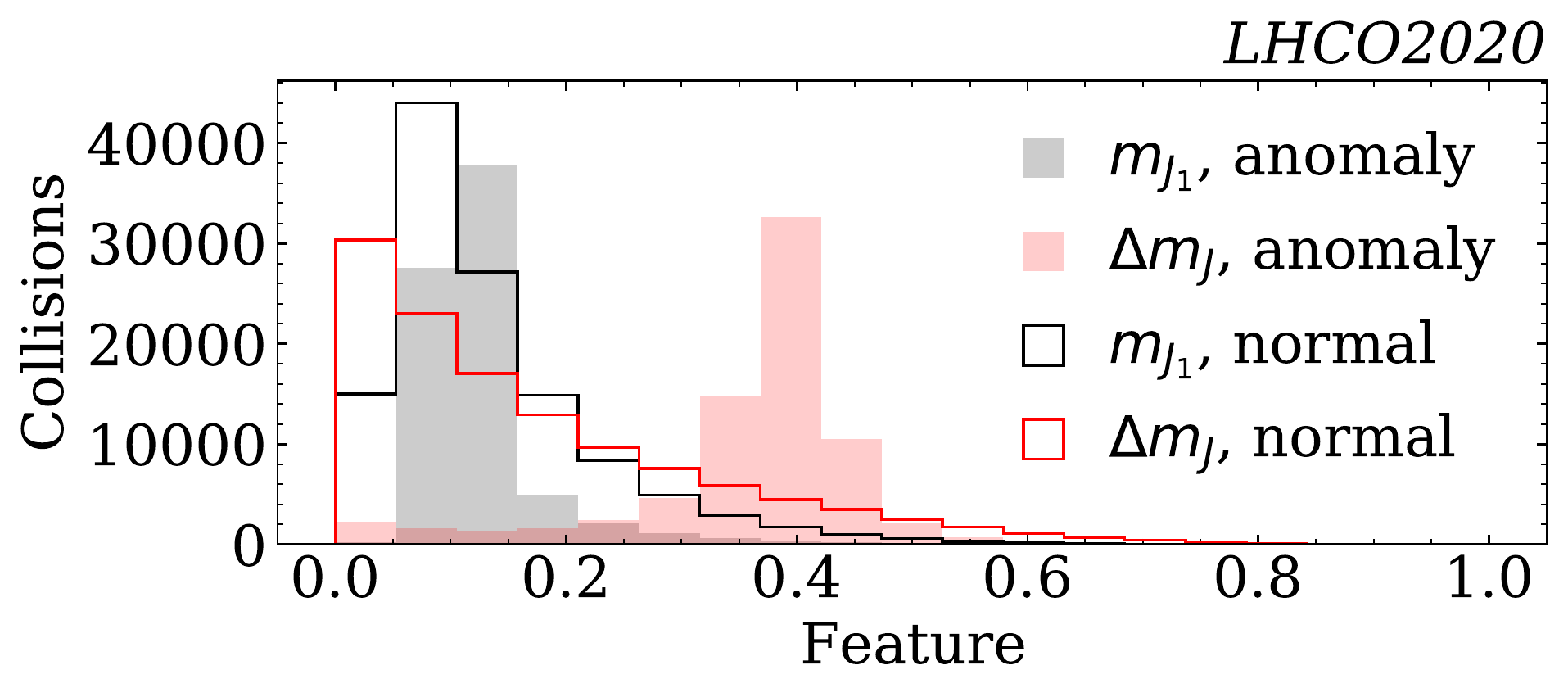}
\includegraphics[width=0.47\columnwidth]{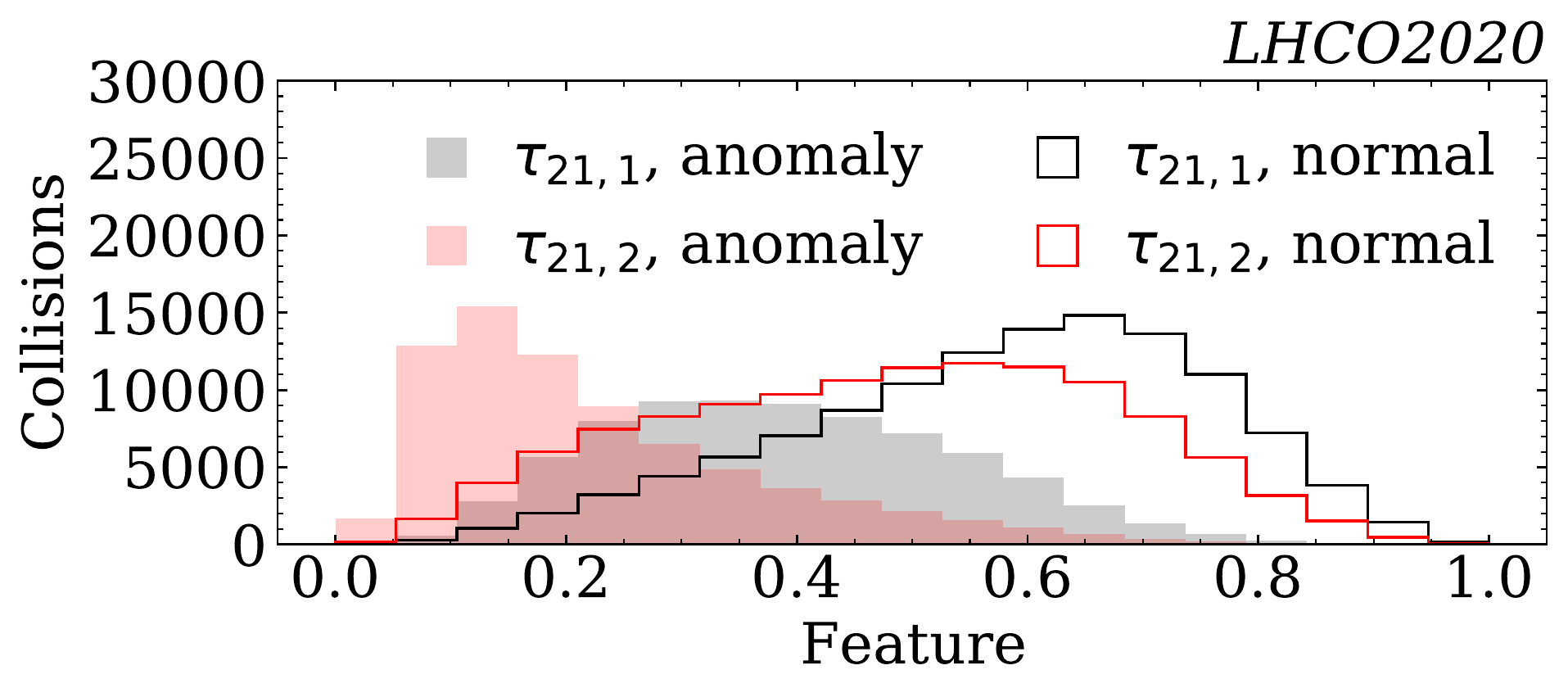}
\caption{Histograms of the four high-level features provided in the LHCO2020 data. The features in the right plot are dimensionless and the features in the left plot are given in units of TeV.}
\label{fig:features}
\end{figure}

\begin{figure}[h!]
\includegraphics[width=0.4\columnwidth]{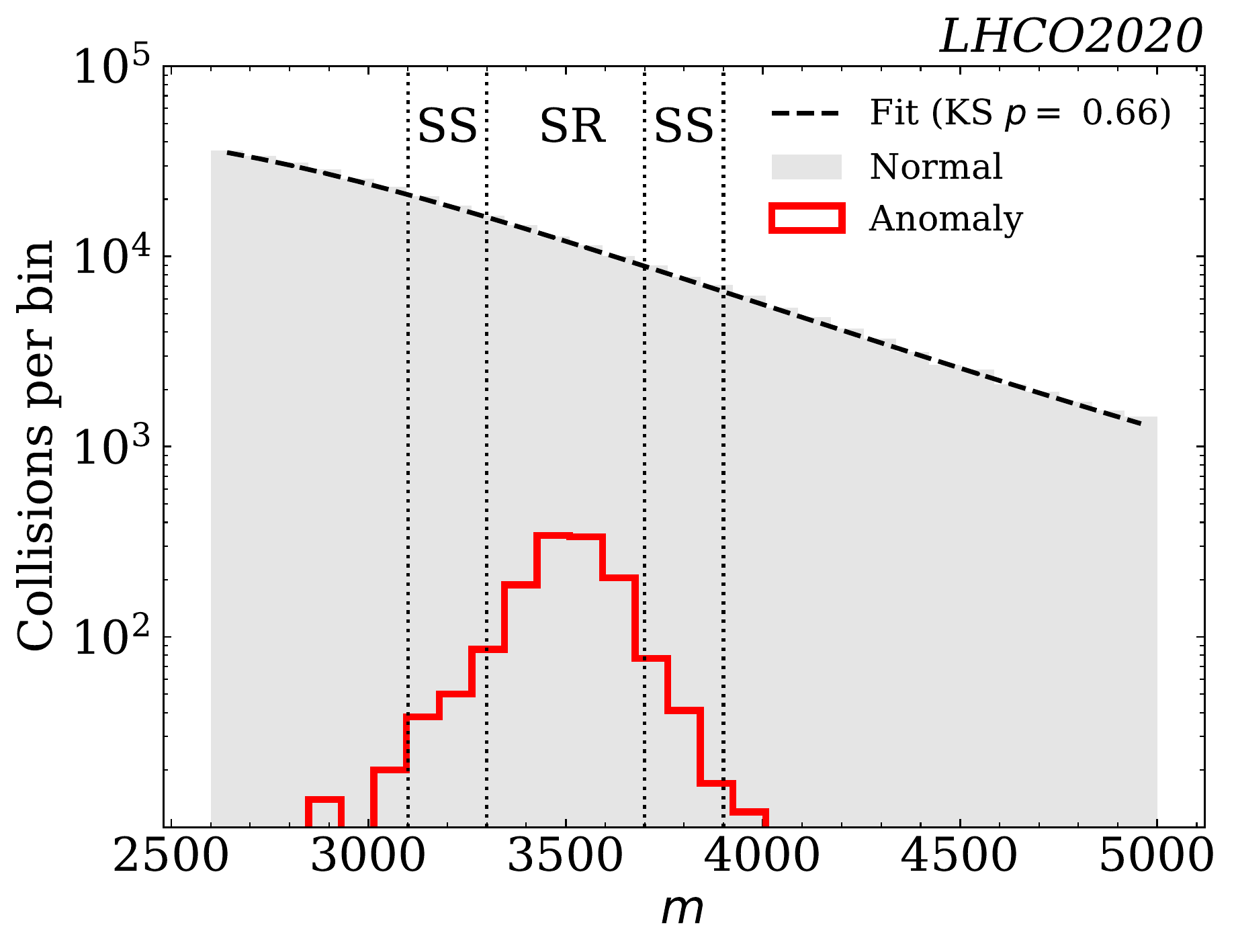}
\caption{A histogram of the resonant feature $m$ in units of GeV with a parametric fit ($\alpha_0(1-m)^{\alpha_1}m^{\alpha_2+\alpha_3\log(m)}$) using the SB data overlaid.  The fit Kolmogorov-Smirnov (KS) $p$-value is well above 0.05 in the SB.}
\label{fig:mjj}
\end{figure}

Many approaches in the LHC Olympics challenge were based on these features,
instead of the low-level features. Plots of these high-level (histograms marginalized over the rest of the feature space) are shown in Fig.~\ref{fig:features}. We see that many of them are quite useful in separating signal vs background. The resonant feature is shown in Fig.~\ref{fig:mjj}.

\subsection{Black Box 1}
\label{sec:challenge_bb1}

This box contained the same signal topology as the R\&D dataset (see Fig.~\ref{fig:bb1sig}) but with different parameters for the anomalous particles, in order that a method trained exclusively on the R\&D dataset could not trivially succeed on the Black Box dataset. A total of 834 signal events were included (out of a total of 1M events in all).  This number was chosen so that the approximate local significance inclusively is not significant.\footnote{It is important to keep in mind that in particle physics, the discovery threshold is conventionally taken to be $5\sigma$, corresponding to a $p$-value of $3\times10^{-7}$ under the null hypothesis.} In order to emulate reality, the background events in Black Box 1 are different to the ones from the R\&D dataset.  The background still uses the same generators as for the R\&D dataset, but a number of \textsc{Pythia} and \textsc{Delphes} settings were changed from their defaults to mimic the domain
shift between simulation and experimental data.

\subsection{Black Box 2}
\label{sec:challenge_bb2}

This sample of 1M events was background only. The background was produced using a different publicly-available and standard particle-physics event generation tool, \textsc{Herwig++}~\cite{Bahr:2008pv}, instead of \textsc{Pythia}. Also, it used a modified \textsc{Delphes} detector card that is different from Black Box 1 but with similar modifications on top of the R\&D dataset card. 

\subsection{Black Box 3}
\label{sec:challenge_bb3}

The signal was based on Ref.~\cite{Agashe:2016rle,Agashe:2016kfr} and consisted of a hypothetical heavy BSM particle
with two different decay modes resulting in
two collimated showers of particles (``dijets")
or
with three collimated showers of particles (``trijets")
as illustrated in Fig.~\ref{fig:bb1sig} center and right. 
These signals are inspired by theories introducing extra dimensions of space-time.
1200 dijet events and 2000 trijet events were included along with Standard Model backgrounds in Black Box 3 (for a total of 1M events).  These numbers were chosen so that an analysis that found only one of the two modes would not observe a significant excess.  The background events were produced with modified \textsc{Pythia} and \textsc{Delphes} settings (different than the R\&D and other Black Box datasets).

\subsection{Evaluation of the Challenge}

During the initial challenge phase (see~\cite{2101.08320}), only the signal 
contained in the R\&D Dataset was known to participants. For this, both the 
physical properties (decay topology, masses) and per-event labels were given. 
No such information was made available for Black Box 1--3. 
Participants were asked to submit (separately for each Black Box): 
I) A p-value associated with the dataset having no new particles (null hypothesis);
II) As complete a characterization of the new physics as possible (in text-form)
(e.g. masses and decay modes of all new particles with associated uncertainties);
and III) How many signal events (central value and uncertainty) are in the dataset (before any selection criteria).

After the challenge phase, the physical properties and datasets with added per-event labels (signal or background) were made public, rendering the initial evaluation criteria obsolete. However, as better signal identification will aid better anomaly detection, 
quantities such as accuracy, area under the curve (AUC), or significance improvement (SIC, defined as the ratio of true positive rate over the square root of the false positive rate for a given working point) are useful metrics to report. We stress that while these metrics utilize the truth labels (signal or background) during the evaluation stage, a successful anomaly detection method would ideally not use these labels during the training stage.

\section{Smooth background anomaly detection}
\label{sec:methods}

Without a particular anomaly hypothesis, it is not possible to construct the optimal~\cite{neyman1933ix} classifier $p(x|\mathrm{anomaly})/p(x|\mathrm{normal})$. However, one can construct a related test statistic that strives to achieve optimality for a related hypothesis test: is the data in the SR more consistent with (a model of) itself or with a prediction for the normal data in the SR?  In this case, the optimal test statistic would be $\mathcal{L}=p(x|\mathrm{data},m\in\mathrm{ SR})/p(x|\mathrm{normal},m\in\mathrm{SR})$.  Rejecting the null hypothesis would be evidence of an anomaly.  We describe two approaches that exploit the features of the resonance group anomaly detection in order to approximate this test statistic directly from data.  

\paragraph{Classification Without Labels (CWoLa)~\cite{Metodiev:2017vrx,Collins:2018epr,Collins:2019jip}:} Due to the smoothness condition, $p(x||m-m_0|<\delta,\mathrm{normal})\approx p(x|\delta <|m-m_0|<\epsilon,\mathrm{normal})$, where $\epsilon > \delta$.  We call $\delta <|m-m_0|<\epsilon$ the short sideband (SS).  In the CWoLa protocol, one trains a classifier using $x$ (without $m$) to distinguish data from the SR from data in the SS.  The values of $\epsilon$ and $\delta$ are chosen to have enough examples in both regions, but also to make the regions as close in $m$ as possible.  The observation of Refs.~\cite{Metodiev:2017vrx,Collins:2018epr,Collins:2019jip} (see also the analog with label noise in Ref.~\cite{blanchard2016classification}) is that the classifier trained in this way is monotonically related to $\mathcal{L}$ when optimal.  Any classifier can be used and can be chosen based on the details of the data.  An advantage of the CWoLa approach is that the problem of density estimation is converted into classification, a comparably easier problem.  However, this approach relies strongly on the smoothness assumption and an additional assumption about the feature space: {\it that the CWoLa classifier cannot learn $m$ from $x$.}


\paragraph{Anomaly Detection with Density Estimation (ANODE)~\cite{Nachman:2020lpy}:} The smoothness and resonance conditions can also be used to estimate $p(x|m,\mathrm{normal})$ directly from the SB.  Then, this density can be interpolated into the SR.  One can also estimate the probability density $p(x||m-m_0|<\delta)$ directly in the SR.  The ratio of the direct and interpolated densities will approximate $\mathcal{L}$ when optimal.  Any explicit conditional density estimation strategy will work; the authors of Ref.~\cite{Nachman:2020lpy} used a masked auto-regressive~\cite{NIPS2017_6828} normalizing flow~\cite{pmlr-v37-rezende15}.  An advantage of ANODE is that $p(x|\mathrm{normal})$ can be different in the SR and SB; as long as it is smooth enough so that one can use the interpolation power of neural networks to estimate the density in the SR from the SB, the procedure should work.

For both CWoLa and ANODE, the $\mathcal{L}$ estimate is used to enhance the presence of a potential anomaly.  As we do not know ahead of time how many anomalous examples there may be, we make a small number of fixed choices $\mathcal{L} > \lambda$, where $\lambda$ could be defined by the fraction of normal events that pass the selection.  After this requirement, we need to estimate $\Pr(\mathcal{L}>\lambda|\mathrm{normal})$ to compute a $p$-value in the SR for the observed data.  This tail probability can be estimated by once again using the SB. Now, we simply need to estimate the expected number of normal examples that would pass $\mathcal{L}>\lambda$ in the SR.  The probability mass function is then Poisson with this mean.  The average value can be estimated using a histogram in $m$ and interpolating from the SB (see Fig.~\ref{fig:mjj}).

\section{Experiments}
\label{sec:experiments}

State of the art performance in on this problem --- as measured by most closely predicting properties of an unseen anomaly in a blind study on the first Black Box --- is achieved by density estimation~\cite{2101.08320}.
However, no correct identification of the anomaly was claimed for the more challenging third Black Box during the blind phase, leaving such more-complex multi-group anomalies as an open problem.

\begin{figure}[h!]
\includegraphics[width=0.4\columnwidth]{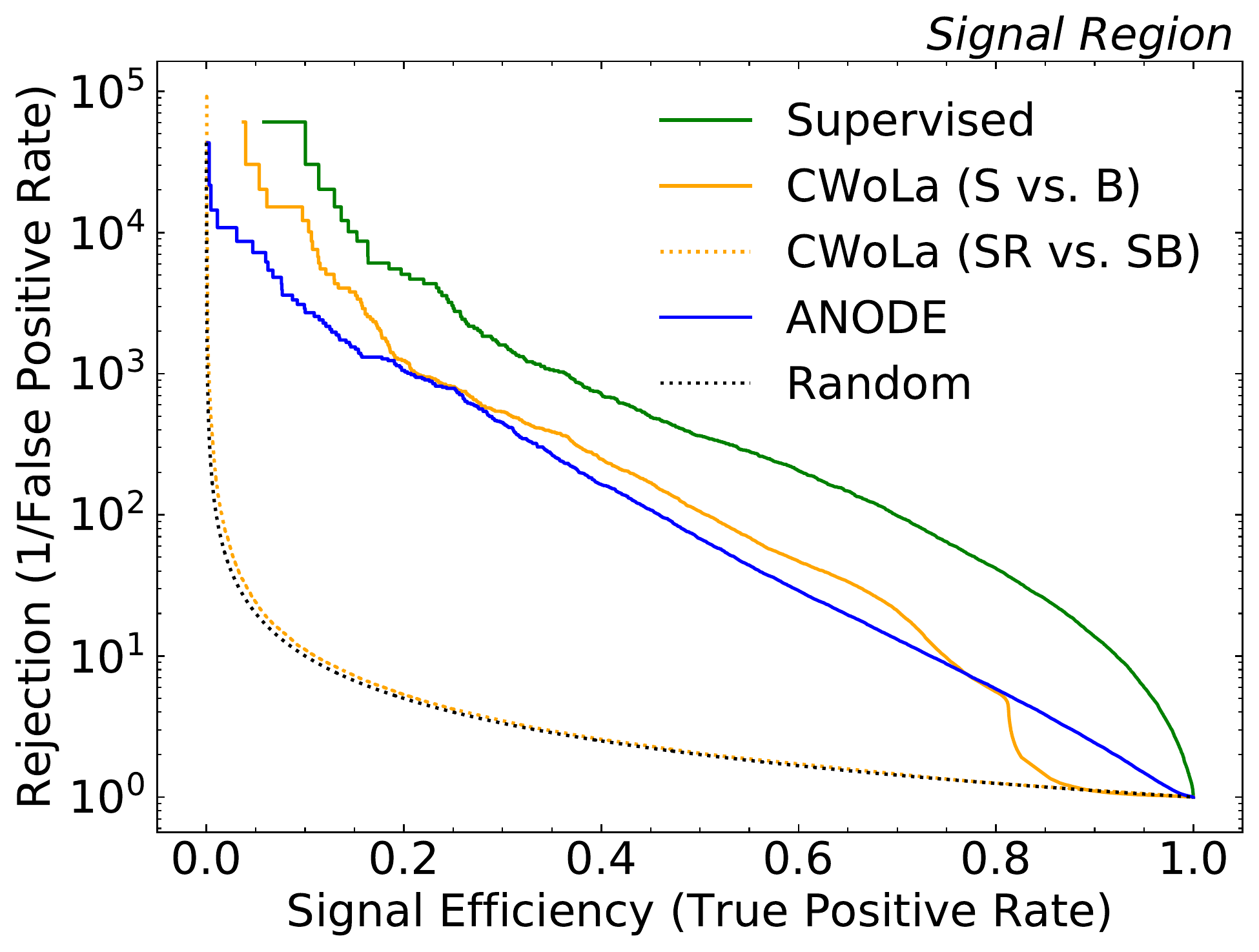}
\hspace{1cm}
\includegraphics[width=0.4\columnwidth]{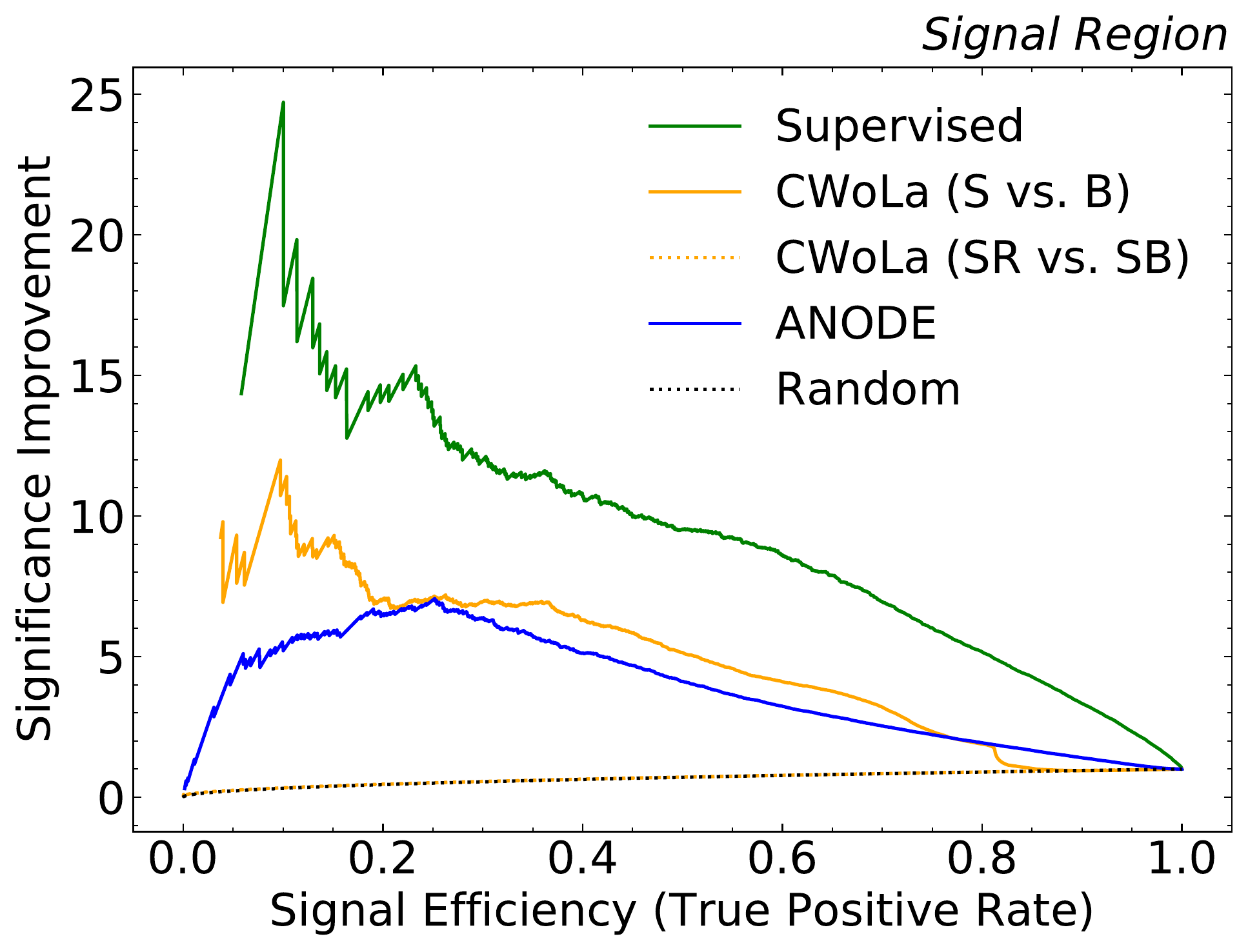}
\caption{ROC curve (left) and significance improvement curve (right) for different anomaly detection algorithms. A detailed explanation of the different lines is provided in the text. Reproduced from Ref.~\cite{Nachman:2020lpy}.}
\label{fig:roc}
\end{figure}

To provide quantitative results, we reproduce in Fig.~\ref{fig:roc} (top) the Receiver Operation Characteristic (ROC) curve for several methods on the R\&D dataset. All algorithms use the high-level observables $(m_{j_1}, \Delta m_j, \tau_{21,1},\tau_{21,2})$ introduced in Sec.~\ref{sec:challenge_rnd} as their input.
Shown are: 
a classifier assuming known anomaly/background labels (Supervised), CWoLa's performance on the SR/SB classification task, CWoLa's performance on the anomaly detection task (S vs.~B), density estimation (ANODE),
and random guessing (Random) for reference. As can be seen from the figure, CWoLa is nearly random when trying to classify SR from SB; this is the expected (and desired) behavior indicating that the central assumption of CWoLa -- that $p(x|\mathrm{normal},m\in\mathrm{SR})\approx p(x|\mathrm{normal},m\in\mathrm{SB})$ -- is satisfied for this feature set. 
Maximal performance --- highest area under the curve (AUC) --- is of course achieved by supervised training.  However, the less-than-supervised approaches still have excellent sensitivity. 


In Fig.~\ref{fig:roc} (bottom) the possible gain in significance achieved by selecting a working point of a given signal efficiency is shown. This gain is calculated as the ratio of true positive rate and the square root of the false positive rate. It is an estimate of the significance assuming uncertainties dominated by Poisson statistics.
Most relevant are the maxima of these curves, which reach $7$ for ANODE (for a TPR of 0.25) and more than $10$ for CwoLa (for a TPR of 0.1). Put differently, by selecting examples based on a working point chosen for ANODE (CWoLa) corresponding to a true positive rate of 0.25 (0.1), the statistical significance of the signal is improved seven-fold (more than ten-fold). The supervised approaches reach an even higher maximum above twenty, but rely on perfect truth labels, which are of course never present in real data. 

\section{Conclusion}

Unsupervised group anomaly detection is an active and intensely studied topic in fundamental physics where the goal is to detect faint signals hinting at new forces of Nature. Our contribution outlines the key assumptions of this challenge to facilitate significant contribution from non-domain experts. These assumptions can be applied to other domains as well and yield a potentially novel perspective on a wide range of group anomaly detection problems. This is further aided by a curated and validated challenge dataset following FAIR principles --- thereby fulfilling a well documented community need.

We review two well performing algorithms on the LHCO2020 datasets and show how robust anomaly detection without supervision and without group labels is possible. These results and implementation are available as reference for future studies\footnote{\url{https://github.com/davidshih17/ANODE}, 
\url{https://github.com/Jackadsa/CWoLa-Hunting/tree/tf2/LHCO-code}}. Both the classification-based CWoLa and the density-based ANODE methods are shown to increase an approximation of the potential improvement in statistical significance by a factor of five or more.  This large effect demonstrates the crucial power of group anomaly detection to increase the sensitivity of fundamental physics experiments, and heralds the promise of these methods for other domains.

Together, this contribution provides a bridge between the fundamental science and machine learning communities and introduces the next big challenge in particle physics (following the discovery of the Higgs boson) to a new audience.

\begin{acks}
The authors thank the participants of the LHC Olympics for many interesting discussions on using anomaly detection in particle physics.
BN and GK are grateful to the NHETC Visitor Program at Rutgers University for the generous support and hospitality during the spring of 2019 where the idea for the LHC Olympics 2020 was conceived.
GK acknowledges support by the Deutsche Forschungsgemeinschaft (DFG, German Re\-search Foundation) under Germany’s Excellence Strategy – EXC 2121 ``Quantum Universe'' – 390833306. BN was supported by the Department of Energy, Office of Science under contract number DE- AC02-05CH11231. DS was supported by the DOE under Award Number DOE-SC0010008.

\end{acks}

\appendix

\section{Glossary of physics acronyms used in the paper}

\begin{table}[h!]
  \begin{center}
    \caption{A glossary of the physics acronyms used in this work.}
    \label{tab:table1}
    \begin{tabular}{c | c} 
      Acronym & Definition\\
      \hline
    ANODE & Anomaly Detection with Density Estimation\\
    BSM & Beyond the Standard Model\\
    CWoLa & Classification without Labels\\
      LHC & Large Hadron Collider \\
    LHCO & LHC Olympics \\
    SB & Sideband (region)\\
    SR & Signal or Search Region\\
    SS & Short Sideband\\
    \end{tabular}
  \end{center}
\end{table}

\bibliography{refs}

\end{document}